\documentclass[sigconf,screen,authorversion,nonacm]{acmart}
\usepackage{fancyhdr}
\AtBeginDocument{%
    \addtolength{\footskip}{2.0\baselineskip}%
    \fancyfoot[L]{\textit{\textbf{Preprint --- do not distribute.}}}%
}

\AtBeginDocument{%
  \providecommand\BibTeX{{%
    \normalfont B\kern-0.5em{\scshape i\kern-0.25em b}\kern-0.8em\TeX}}}

\usepackage[ruled,vlined]{algorithm2e}
\usepackage{gensymb}
\SetKwComment{Comment}{/* }{ */}

\setcopyright{acmcopyright}
\copyrightyear{2025}
\acmYear{2025}
\acmDOI{XXXXXXX.XXXXXXX}

\acmConference[KDD '25]{Conference on Knowledge Discovery and Data Mining}{August 3-7, 2024}{Toronto, Canada}
\acmISBN{978-1-4503-XXXX-X/18/06}

\begin{document}

\title[Budgeted Online Active Learning with Expert Advice and Episodic Priors]{Budgeted Online Active Learning \\ with Expert Advice and Episodic Priors}

\author{Kristen Goebel}
\email{goebelk@oregonstate.edu}
\affiliation{%
  \institution{Oregon State University}
  \city{Corvallis}
  \state{Oregon}
  \country{USA}
}

\author{William Solow}
\email{soloww@oregonstate.edu}
\affiliation{%
  \institution{Oregon State University}
  \city{Corvallis}
  \state{Oregon}
  \country{USA}
}

\author{Paola Pesantez-Cabrera}
\email{p.pesantezcabrera@wsu.edu}
\affiliation{%
  \institution{Washington State University}
  \city{Pullman}
  \state{Washington}
  \country{USA}}

\author{Markus Keller}
\email{mkeller@wsu.edu}
\affiliation{%
  \institution{Washington State University}
  \city{Pullman}
  \state{Washington}
  \country{USA}}

\author{Alan Fern}
\email{afern@oregonstate.edu}
\affiliation{%
  \institution{Oregon State University}
  \city{Corvallis}
  \state{Oregon}
  \country{USA}
}

\renewcommand{\shortauthors}{Goebel, Solow, Pesantez-Cabrera, Keller, and Fern}

\begin{abstract}
This paper introduces a novel approach to budgeted online active learning from finite-horizon data streams with extremely limited labeling budgets. In agricultural applications, such streams might include daily weather data over a growing season, and labels require costly measurements of weather-dependent plant characteristics. Our method integrates two key sources of prior information: a collection of preexisting expert predictors and episodic behavioral knowledge of the experts based on unlabeled data streams. Unlike previous research on online active learning with experts, our work simultaneously considers query budgets, finite horizons, and episodic knowledge, enabling effective learning in applications with severely limited labeling capacity. We demonstrate the utility of our approach through experiments on various prediction problems derived from both a realistic agricultural crop simulator and real-world data from multiple grape cultivars. The results show that our method significantly outperforms baseline expert predictions, uniform query selection, and existing approaches that consider budgets and limited horizons but neglect episodic knowledge, even under highly constrained labeling budgets.

\end{abstract}

\ccsdesc[500]{Computing methodologies~Online learning settings}
\ccsdesc[500]{Computing methodologies~Active learning settings}
\ccsdesc[300]{Applied computing~Agriculture}

\keywords{Online Learning, Active Learning, Limited Budget, Agriculture}


\maketitle

\section{Introduction}

Consider a vineyard manager tasked with predicting the cold hardiness of grape vines during the dormancy period---a property heavily influenced by daily weather patterns. Accurate predictions can guide the application of frost mitigation measures, helping to prevent crop loss due to freezing temperatures. However, obtaining ground-truth cold-hardiness measurements for model learning is both costly and time-consuming, requiring field sampling and laboratory analysis. Typically, only 2--5 measurements can be collected in a season, raising two critical questions: when should these limited samples be collected, and how can we effectively learn from such small sample sizes?

This scenario exemplifies a broader class of problems in \emph{budgeted online active learning (BOAL)}, where we aim to learn a target function through strategic label queries under three key constraints:
\begin{enumerate}
    \item \textbf{Online querying}: Query decisions must be made sequentially as data arrives
    \item \textbf{Finite horizon}: A fixed time window for data collection (e.g., a growing season)
    \item \textbf{Strict query budgets}: A fixed budget for the horizon with a focus on small-budget scenarios with 2--10 labels total
\end{enumerate}
These constraints arise in many real-world application areas such as agricultural monitoring, environmental sensing, industrial process control, and personalized medicine. As discussed in Section~\ref{sec:related}, prior work in active learning fails to address the combined challenges of online querying, finite horizons, and strict budgets---particularly when budgets are extremely small. Our work bridges this gap by introducing a framework that leverages two sources of prior knowledge: expert advice and historical episodic data.

First, we adapt the classic paradigm of online learning from expert advice, providing the learner with a collection of pre-existing predictors that is likely to contain a ``good enough" predictor. This enables rapid learning by identifying effective experts early, even with minimal labels. While prior work has studied active learning for the experts setting, no work to our knowledge, has considered both budget and finite-horizon constraints. 

Second, we introduce novel query selection methods that leverage historical episodes of the input stream (e.g., decades of weather data). In particular, we draw on recent progress in the area of prophet-inequality problems as well as develop a simple, but effective, adaptive approach. Intuitively, the prior data can help inform the fundamental decision of whether to sample at the current time, or wait for a potentially better time in the future. To the best of our knowledge, this is the first work that has used such prior episodic data to inform query selecting in BOAL. 

We provide experimental results on a variety of BOAL problems derived from a realistic multi-crop agricultural simulator and decades of real-world grape cold-hardiness data. The results demonstrate that our approach is able to significantly improve over baseline predictors with very small sample sizes. We also show that our approach outperforms uniform query selection and existing work that does not use prior historical data to inform query selection. 

In summary, the main contributions of this paper are: 
\begin{enumerate}
    \item A general BOAL framework for extremely small budgets. 
    \item The first integration of the learning-from-experts paradigm into BOAL. 
    \item The first integration of prior unlabeled episodic data into online query selection.  
    \item An empirical evaluation on both realistic-simulated and real-world agricultural data, demonstrating the effectiveness of our approach.
\end{enumerate}

\section{Related Work}
\label{sec:related}

The majority of prior work on active learning follows a pool-based paradigm, where a set of unlabeled data is available to freely select from for labeling. However, early work considered an online active-learning setting \cite{atlas1990}, leading to algorithms such as query-by-committee for various types of hypothesis classes \cite{seung1992,freund1997a}. Even so, to the best of our knowledge, there is no prior work that fully captures all of our problem features. Below, we overview the most closely related categories of prior work. 

{\bf Online Active Model Selection.} The most closely related work to our expert advice setting is pool-based model-selection (e.g. \cite{hara2024}), where a finite set of models is available to select among. However, it is unclear how to adapt these pool-based methods to our online problem. There has also been work in online active model selection (e.g. \cite{karimi2021,liu2022}), but, while these works aim to minimize the number of label queries, they do not explicitly reason about finite budget and horizon constraints, which are required for our problem setting. In addition, these approaches are specialized for classification tasks and are not directly adaptable to real-valued labels, which are essential for our motivating applications. 

{\bf Active Learning with Expert Advice.} Online learning with expert advice has been the focus of significant prior work \cite{littlestone1994,freund1997b,cesa1997,herbster1998}, where a finite set of experts makes online predictions with labels being revealed immediately after each prediction. More recently, active learning variants have been studied, where the learning algorithm aims to minimize labeling effort by selectively requesting labels of instances as they arrive. For example, one type of approach selects a fixed probability of requesting labels and analyzes the regret and query requirements \cite{cesa2005,mitra2020}. Other approaches use more selective query conditions, which depend on the prior labeling history \cite{zhao2013,hao2018,kumar2022,castro2024}, however, these approaches are specialized for classification tasks and not directly adaptable to our real-valued labels setting with exceptions being \cite{truong2021}. None of these approaches explicitly reason about fixed query budgets over finite horizons. Rather, they focus on reducing the number of queries without reference to particular budget or horizon constraints. Thus, they are not easily adapted to our problem setting. 

{\bf Budget Aware Online Active Learning.} Prior work, such as Zhang et al. \cite{zhang2018}, combine online active learning with query budgets but ignore finite horizons, a critical constraint in time-sensitive applications like agricultural monitoring. More closely related to our setting is prior work on online sample selection along robot trajectories \cite{das2015,luo2017}, which reasons about both finite-horizons and budget constraints. The approach is based on applying the multi-choice submodular secretary algorithm \cite{bateni2013} to a submodular active-learning objective. However, the algorithm is non-adaptive, in the sense that the observed labels from prior queries are not used to inform future query selections. An adaptive variant of this approach extends on adaptive submodular optimization \cite{fujii2016}, which is evaluated on synthetic benchmarks with only i.i.d. data streams. 

None of these prior works consider the incorporation of prior unlabeled data streams to help inform online query selection. Rather they rely on the online secretary framework for label selection, which assumes no knowledge of the stream statistics. Rather, we leverage such prior information by adopting the online selection framework of prophet inequalities and a newly proposed approach for empirical threshold selection. We demonstrate that this can lead to benefits under tight query budgets, with evaluations on realistic problems involving non-i.i.d. streams. 

\section{Budgeted Online Active Learning}
\label{sec:formulation}

We consider \emph{budgeted online active learning (BOAL)} over an online finite-horizon data stream $x_{1:T}=(x_1, x_2, \ldots, x_T)$, where $T$ is the known time horizon and each $x_t \in \mathbb{R}^d$ is a d-dimensional feature vector. Our goal is to approximate a target function $f: x_{1:t}\rightarrow \mathbb{R}$, which maps a history $x_{1:t}=(x_1, \ldots, x_t)$ to a target value. For example, in agriculture, each $x_t$ may correspond to weather station measurements on day $t$, $T$ is the number of days in the growing season, and $f$ corresponds to a plant property that depends on the weather history. 

Our framework addresses the challenge of learning $f$ from the data stream via selective label queries, subject to strict budgetary and temporal constraints. In particular, we focus on applications with a severely limited query budget $B$ (e.g. 2 to 5 queries) where $B \ll T$ and with no ability to query past time points. This emphasis is motivated by real-world scenarios, such as agricultural applications, where each query involves costly field sampling and laboratory analysis. The online active learning protocol unfolds over the finite horizon, during which the algorithm must strategically allocate its query budget as follows:
\begin{enumerate}
\item At each time step $t \in {1, \ldots, T}$, the algorithm observes input $x_t$.
\item The algorithm decides whether to query for label $y_t$, subject to the constraint that the total queries cannot exceed $B$.
\item If queried, $y_t$ reveals the value of $f(x_{1:t})$.
\end{enumerate}

This learning setting presents a complex challenge arising from the interplay of three key factors: the online nature of the problem, the finite horizon, and the extremely limited query budget. Unlike batch active learning, our online setting imposes a critical temporal constraint: if the algorithm doesn't query at time $t$, it permanently forfeits obtaining $f(x_{1:t})$. This constraint is intensified by the limited horizon and query budget, yielding a fundamental choice at each time step: \emph{Given the current observation history, should the algorithm expend a query from the limited budget to gain information now, or wait in anticipation of potentially more informative query opportunities before the horizon expires?} Our main contribution is a novel approach to this problem that aims to leverage prior knowledge to address extreme budget constraints. 

\section{Generic Approach}
\label{sec:generic}


In this section, we present the generic algorithm underlying our approach to BOAL. In Sections \ref{sec:experts} and \ref{sec:prophet}, we detail the specific components of our novel instantiation that leverages expert advice and episodic priors.

A typical pool-based active learning algorithm \cite{settles2009} is given a full set of unlabeled examples and iteratively selects an instance to be labeled, immediately updating the prediction model thereafter. Instance selection is commonly done by computing instance scores that capture how much the model might improve if updated based on the observed label. For example, a common instance score is some measure of the current model's uncertainty about the label, as highly uncertain instances often provide the most informative updates. The fundamental challenge in the online setting, in contrast to pool-based approaches, is that it is not possible to maximize over the scores of all instances. Rather, the decision of whether to query the current instance now or never can only be based on the score of the current and previous instances. We will refer to this problem of trying to select a single instance of maximum score from a data stream as \textsc{OnlineMax}.

When all that is known about the stream is that a set of unknown scores arrive in random order, the classic \emph{secretary algorithm} \cite{dynkin1963} is known to be optimal. This algorithm solves \text{OnlineMax} as follows when applied to a sub-sequence of scores $(s_{t_0},\ldots,s_{t_e})$:
\begin{enumerate}
    \item Observe the first $t' = \lfloor (t_e-t_0+1)/e \rfloor$ scores.
    \item $S_{\max}=\max\{s_1,\ldots,s_{t'}\}$ 
    \item For $t\in \{t'+1,\ldots, T\}$\\ if $\left(s_t > S_{\max} \text{  or } t=t_e\right)$ then select $t$ and stop. 
\end{enumerate}
Despite its simplicity, this strategy achieves the best possible approximation ratio of $1/e$ in expectation relative to the best value in hindsight. More closely related to online budgeted active learning, this classic problem has been extended to the \emph{submodular secretary problem} \cite{bateni2013} where $B$ instances are selected from the stream to maximize the value of a submodular set function. The approximation algorithm for this variant sequentially applies the classic secretary algorithm $B$ times to evenly divided segments of the data stream. This algorithm achieves an approximation ratio of $(1-1/e)/7$ in expectation for the standard random-arrival stream model. 

\begin{algorithm}[t]
\caption{Generic Budgeted Online Active Learning}
\label{alg:general}
\KwIn{Time horizon $T$, query budget $B$, data stream $x_{1:T}$}
\KwOut{Updated prediction model $\hat{f}$}
\SetKwProg{Fn}{Function}{:}{}

\textbf{Requires:} \textsc{OnlineMax} (Sec. \ref{sec:prophet}); \textsc{Update}; \textsc{Score} (Sec. \ref{sec:experts})\\

Initialize prediction model $\hat{f}$

\For{$i \gets 1$ \KwTo $B$}{
    Initialize \textsc{OnlineMax}\;
    \For{$t \gets \lfloor\frac{T}{B}\rfloor(i-1)+1$ \KwTo $\lfloor\frac{T}{B}\rfloor i$}{
        \If{\textsc{OnlineMax}(\textsc{Score}($x_t$)) $=$ select \textbf{or} $t = \lfloor\frac{T}{B}\rfloor i$}{
            $y_t \gets \textsc{Query}(x_t)$\\
            \textsc{Update}($\hat{f}$, $(x_t, y_t)$)
            \textbf{break}\;
        }
    }
}
\Return{$\hat{f}$}
\end{algorithm}

Our approach and those of prior work on BOAL \cite{fujii2016,luo2017} follow the structure of the submodular secretary algorithm. The generic algorithm, shown in Algorithm \ref{alg:general}, begins by initializing the prediction model $\hat{f}$ and then processes the stream in $B$ segments, each of size $T/B$ (for simplicity, we assume $T$ is a multiple of $B$). For each segment, an algorithm for \textsc{OnlineMax} is applied to the scores assigned to instances by an active learning score function \textsc{Score}, which can depend on the current model and any other prior stream information. The selected instance in each segment is used to update the model before proceeding to the next segment, allowing the algorithm to adapt to new label information.

The prior works \cite{luo2017,fujii2016} are instances of Algorithm 1, with a notable difference in Luo et al.'s approach of updating the model only once after all $B$ instances have been queried. Both studies learn models from scratch using Gaussian Process and linear SVM respectively, employing standard uncertainty-based measures for \textsc{Score}. These model choices are impractical for extremely small sample sizes. Our work addresses this limitation through the use of expert advice (Section \ref{sec:experts}), enabling effective learning even with severely constrained budgets. We also note that these prior works have included limited empirical evaluations. Luo et al. focus on selecting samples along a pre-defined robot trajectory, while Fujii and Kashima provide experiments on several synthetically generated streams derived from i.i.d. classification benchmarks.

A more significant distinction between prior work and ours lies in the choice of \textsc{OnlineMax}. Previous works rely on the classic secretary algorithm, which 
makes no use of prior knowledge about data stream statistics. This can lead to suboptimal sample selection. For example, consider a case where we can estimate the distribution of the expected maximum value for each data stream segment. The secretary algorithm might select an instance with a score far below this estimate, potentially wasting a valuable query. Our key innovation, detailed in Section \ref{sec:prophet}, is to move beyond this uninformed framework and use methods that can leverage statistics from prior stream data, allowing for more informed and efficient query decisions.

\section{Leveraging Expert Advice}
\label{sec:experts}

To address the challenge of extremely limited budgets, we adopt the classic online learning framework of learning from expert advice \cite{cesa1997}. This approach leverages a collection of expert predictors, enabling effective learning from very small sample sizes by quickly identifying a good mixture of experts. Below, we describe the framework and our particular instantiations for the \textsc{Update} and \textsc{Score} sub-routines required by Algorithm \ref{alg:general}.

\subsection{Classic Framework} 

The framework formulates online prediction along a data stream $x_{1:T}=(x_1, x_2, \ldots, x_T)$ with corresponding labels $y^*_t\in \mathcal{Y}$, making no assumptions about the stream or label characteristics. The learner is given a set of $N$ expert predictors $F=\{f_1,f_2,\ldots,f_N\}$, where $f_i : x_{1:t} \rightarrow \mathcal{Y}$. No assumptions are made about the experts, but in a typical application, $F$ is selected to likely contain at least one ``good enough" expert. At each step $t$ the learner: 
\begin{enumerate}
    \item Observes $x_t$ 
    \item Selects an expert $f_i\in F$ and predicts $y_t=f_i(x_{1:t})$
    \item Observes prediction loss $l_{i,t} = l(f_i(x_{1:t}),y^*_t)$ of each expert $i\in \{1,\ldots,N\}$
\end{enumerate} 
The cumulative loss of the learner after $T$ steps is $L_T = \sum_{t=1}^T l(y_t,y^*_t)$ and of each expert is $L_{i,T}=\sum_{t=1}^T l_{i,t}$. The learning objective is to minimize the expected cumulative regret, which is the difference between $L_T$ and best expert's regret $\min_i L_{i,T}$. 

A simple, yet powerful, algorithm for this framework is \textsc{Hedge} \cite{freund1997b}, which computes a weight $w_{j,t}$ for each expert $j$ at each time step $t$. The weights are initialized to a uniform value of 1. At each step $t$, the learner makes a prediction by using the current weights to compute a distribution over experts $$p_{i,t} = \frac{w_{i,t}}{\sum_{j=1}^N w_{j,t}}$$ and drawing an expert according to this distribution. Note that in our experiments, which involve real-valued labels, instead of sampling an expert at each step we return the weighted average prediction given by 
\begin{equation}
\label{eq:output}
    y_t = \sum_i p_{i,t}\cdot f_i(x_{1:t}).
\end{equation}

After observing the expert losses, the next weights are computed using a multiplicative update rule: 
\begin{equation}
\label{eq:hedge}
w_{i,t+1} = w_{i,t} \cdot \exp(-\eta \cdot l_{i,t}),    
\end{equation} 
where $\eta$ is a learning rate. This update rule exponentially decreases the weight of experts that perform poorly. Intuitively, experts with lower cumulative losses will be assigned higher probabilities in future steps. Remarkably, with an appropriate choice of $\eta$, this algorithm achieves an expected cumulative regret that grows as $O(\sqrt{T \log N})$. This sublinear regret bound ensures that the algorithm's performance approaches that of the best expert as the sequence length increases.

\subsection{Active Framework} 

Our online active learning setting differs from the classic framework as labels are only revealed when a query is issued, not at every time step. We adapt the \textsc{Hedge} algorithm for the \textsc{Update} subroutine in Algorithm \ref{alg:general} as follows: If a query is issued at time $t$, we observe the expert losses and perform the usual weight update from Equation \ref{eq:hedge}. On time steps with no query, we set the losses of all experts to zero, resulting in no change to the weights.

Algorithm \ref{alg:general} also requires a \textsc{Score} sub-routine for \textsc{Hedge}. Since our motivating agricultural applications typically involve real-valued labels, we require a \textsc{Score} instantiation appropriate for regression. As reviewed in Section \ref{sec:related}, to our knowledge, only one prior work \cite{truong2021} addresses active learning with expert advice for regression problems. In that work, \textsc{Score} for a time step $t$ is defined as the maximum absolute difference between any two expert predictions, i.e., $\max_{i,j} | f_i(x_{1:t}) - f_j(x_{1:t})|$. While theoretical guarantees were provided for this scoring function, in practice it can be highly uninformative, as it ignores the weights of the experts and a single outlier expert can lead to a high score.

Our scoring function is inspired by the widely-used query-by-committee framework \cite{seung1992}, where an ensemble of predictors is used to compute an active learning score based on disagreement. This approach has been used for regression using the variance of predicted values as the score (e.g. \cite{krogh1994}). In our case, \textsc{Hedge} maintains a committee of weighted experts. Thus, we use weighted variance as our score function: 
\begin{equation}
\label{eq:score}
\text{Score}(x_{1:t})=\sum_i p_{i,t} \left(f_i(x_{1:t}) - y_t\right)^2,  
\end{equation}
where $y_t$ is our prediction (Equation \ref{eq:output}). Intuitively, this function emphasizes instances with more disagreement among highly weighted experts while mitigating the influence of low-weighted experts.

\section{Leveraging Episodic Priors}
\label{sec:prophet}

In many BOAL applications, it's possible to acquire historical data providing sample episodes from the process generating the input stream $x_{1:T}$. For instance, in applications where the input stream corresponds to episodes of weather data, we typically have access to historical weather data over corresponding episodes. This data allows for computing statistics about the expected behavior of our experts (e.g., the mean disagreement score), potentially better informing the selection of query points. However, as described in Section \ref{sec:generic}, prior BOAL work has used the classic secretary algorithm for \textsc{OnlineMax}, which makes minimal assumptions about the input stream and is unable to leverage such prior information.

To incorporate this prior knowledge, below we describe two approaches. Both assume that we have the ability to sample historical stream sub-episodes that correspond to the stream segment of the current call to \textsc{OnlineMax} in Algorithm \ref{alg:general}. In particular, if the call to \textsc{OnlineMax} is on the subsequence $(x_{t_0},\ldots,x_{t_e})$, with start and end times $t_0$ and $t_e$, we generate a sample of $K$ corresponding historical episodes in the same time range denoted by $$E(t_0,t_e) = \left\{x^1_{t_0:t_e}, \ldots, x^K_{t_0:t_e}\right\}.$$ For example, for weather data streams, the sub-episodes correspond to historical weather in the dates ranging from $t_0$ to $t_e$. Importantly, our approaches do not require any labels for this prior data. 

\subsection{Prophet-Secretary Algorithm (PSA)} 

The prophet inequality problem is similar to the secretary problem, but assumes more knowledge of the input stream. The classic version \cite{krengel1977} assumes the stream is constructed from a set of $T$ known independent distributions that can be adversarially ordered to form the stream. A simple stopping rule for this setting can achieve the best possible approximation ratio of 0.5 \cite{samuel1984} relative to a prophet that has perfect hindsight knowledge. More recent extensions assume more knowledge of the stream in order to achieve stronger guarantees. In this work, we consider the prophet-secretary variant \cite{esfandiari2017,ehsani2018}, where the set of $T$ known distributions are encountered in a randomly selected order (i.e. non-adversarial). While this variant doesn't take advantage of correlation structure in a stream, there is an effective algorithm that only require the estimation of a single stream statistic. Rather, variants that require more detailed knowledge, such as the Markov process underlying the stream \cite{Truong2019}, are not practical due to the difficulty of accurately estimating the knowledge (e.g. accurate long-term daily weather). 

PSA \cite{ehsani2018} uses the prior episodes $E(t_0,t_e)$ to compute an estimate of the expected maximum score in an episode denoted by $$\text{OPT} = \frac{1}{K} \sum_{k=1}^K \max_t \{\text{Score}(x^k_t)\}.$$ This estimate converges rapidly according to standard concentration inequalities. Given OPT, the PSA algorithm compares each score in the finite input stream $(s_{t_0},\ldots, s_{t_e})$ to a sequence of decreasing thresholds as follows:
\begin{enumerate}
    \item $\tau_t = \text{OPT}\cdot (1-\exp\left(\frac{t-t_e}{t_e-t_0+1}\right)$
    \item for each $t\in \{t_0,\ldots, t_e\}$ \\ if $\left(s_t > \tau_t\right)$ then select $t$ and stop
\end{enumerate}
PSA achieves the best known approximation ratio of $(1-1/e)$ for the prophet-secretary formulation. Importantly, note that each call to PSA may generate a different sequence of thresholds due to the fact that OPT is computed from the set of prior episodes in $E(t_0,t_e)$, which depends on the specific time interval of the call. 

\subsection{Empirical Threshold Selection (ETS)} 

We propose a simple approach that aims to use the historical data to find a single optimal threshold $\tau$ to use as a stopping rule for \textsc{OnlineMax}. Given the set of historical episodes $E(t_0,t_e)$ and a threshold $\tau$, we can estimate the expected performance of using $\tau$ as a stopping rule as follows: 
\begin{enumerate}
    \item Compute the score function for the current set of experts along each historical episode in $E(t_0,t_e)$.
    \item Let $s_{\tau}^k$ be the first score in $s^k_{t_0:t_e}$ that is larger than $\tau$. 
    \item Return $s^*_{\tau} = \frac{1}{K} \sum_k s_{\tau}^k$ 
\end{enumerate}
The ETS algorithm considers a discrete set of possible thresholds $\mathcal{T} =\{\tau_1,\ldots, \tau_M\}$ and selects the threshold $\tau^*$ with the best estimated performance: $\tau^* = \arg\max_{\tau \in \mathcal{T}} s^*_{\tau}.$ 
Given the input sequence from \textsc{OnlineMax}, ETS then stops at the first score exceeding $\tau^*$.

Like PSA this algorithm can be applied dynamically at each call to \textsc{OnlineMax}, allowing it to adapt to changes in expert weights and score distributions. However, compared to PSA, it more directly attempts to optimize its expected performance relative to the historical data. This comes at the cost of increased computational complexity compared to PSA, as it requires simulating multiple thresholds for all historical episodes at each step.

\section{Experimental Results}
We evaluate the performance of our method against baseline and existing methods. We consider several simulated agricultural problems on various crops and the problem of grape cold-hardiness prediction using real world data, as described below.

\subsection{Agricultural Simulator Problems}


Crop growth models \cite{boote_potential_1996} are traditionally used to model and validate agronomists' understanding of crop processes in the field. In this work, we use the WOFOST crop growth model \cite{van_diepen_wofost_1989}, a high-fidelity model for a wide variety of annual crops subject to nutrient and water fluxes. Ideally, a well-calibrated WOFOST model can act as a digital twin for a particular crop, providing predictions of crop or soil features that are difficult or expensive to obtain. This suggests using BOAL to guide costly data collection in order to calibrate WOFOST models for site-specific agromanagement decisions. 


Our experiments use the WOFOST model to both generate data for simulated BOAL problems and define a set of experts. We use four parameterizations of the WOFOST model \cite{de_wit_25_2019} corresponding to varieties of wheat, maize, sorghum, and millet crops. For all of our results, we run the WOFOST model with historical weather data from 1984 to 2023 at 52$\degree$N, 5$\degree$W provided by the NASA POWER Database. We use an open-loop agromanament policy that alternates biweekly between irrigation and fertilization. 

\begin{figure}[t!]
  \centering
  \includegraphics[width = 0.95\columnwidth]{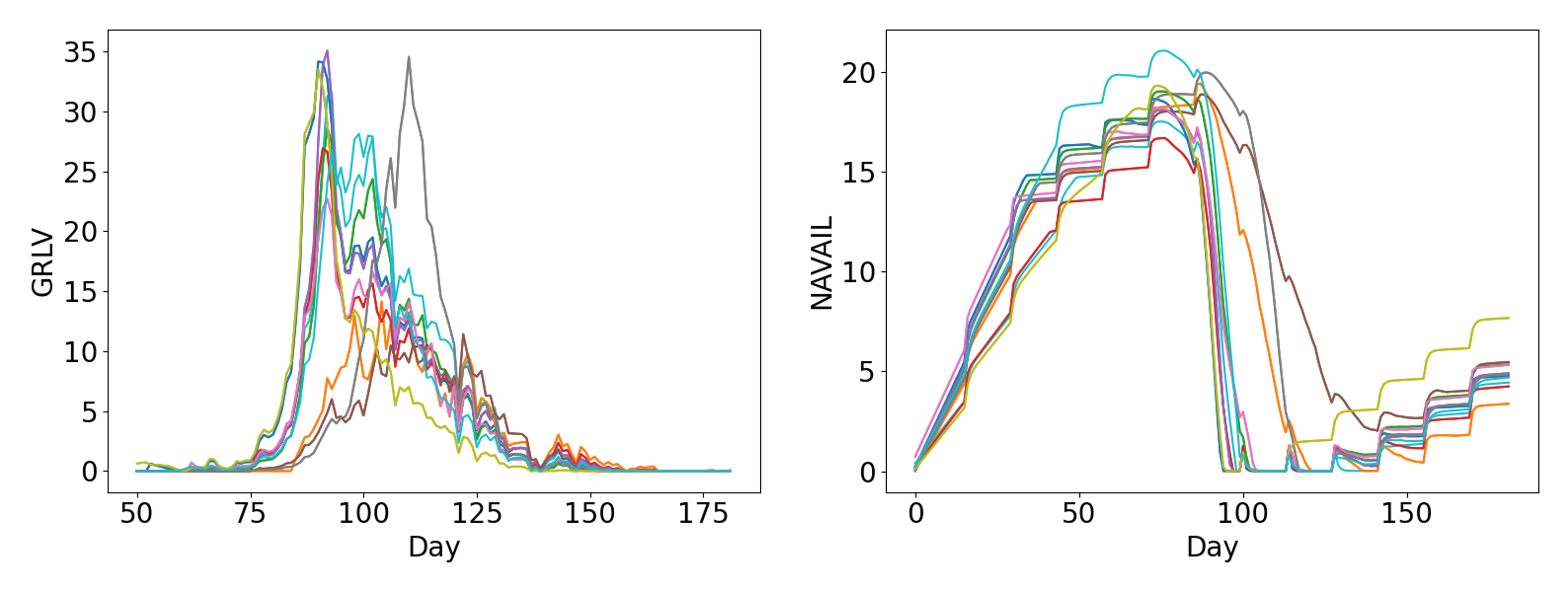}
  \vspace{-1em}
  \caption{Leaf growth rate (GRLV) and available nitrogen (NAVAIL) from set of maize experts for a single season. Each line gives the predictions of a different expert, which are other maize crops with different simulator parameters.}
\label{fig:experts-cropsim}
\end{figure}

For our experimental protocol (Section \ref{sec:protocol}) we generated 15 WOFOST models for each crop by adding uniform noise ($\pm 10\%$) to a subset of crop-specific parameters. These models cover a variety of crop and soil variations that are reasonable to expect in the real world. Figure \ref{fig:experts-cropsim} shows the variety of outputs of the models for a single crop and season of weather data. For our experiments, we consider two different prediction targets from the WOFOST model: 1) the \emph{nitrogen available in the subsoil (NAVAIL)} and 2) the \emph{growth rate of living leaves (GRLV)}. Both of these state features are expensive or time consuming to measure in practice, yet are important for informing agromanagement decisions. 


\begin{figure}[t!]
  \centering
  \includegraphics[width = 0.85\columnwidth]{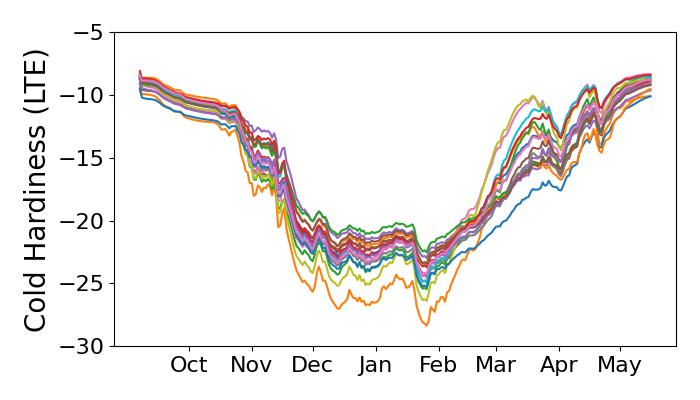}
  \vspace{-1em}
  \caption{Cold-hardiness predictions from set of experts for a single season. Each line gives the predictions of a different expert, which are other grape cultivar models.}
\label{fig:experts-ch}
\end{figure}

\begin{table*}[t!]
  \caption{Comparison of RMSE of label predictions for the different sampling methods. Each method is evaluated on four crops and two labels. The column labels denote 1) the number of samples, 2) the sampling method.}
  \small
  \centering
  \begin{tabular}{llllll|llll|llll|llll}
    \toprule
    Budget & \multicolumn{1}{c}{0} & \multicolumn{4}{c}{2} & \multicolumn{4}{c}{3} & \multicolumn{4}{c}{4} & \multicolumn{4}{c}{10}\\
    \cmidrule(r){2-2}
    \cmidrule(r){3-6}
    \cmidrule(r){7-10}
    \cmidrule(r){11-14}
    \cmidrule(r){15-18}
    Crop, Label & Base & UNI & SA\cite{fujii2016} & PSA & ETS & UNI & SA\cite{fujii2016} & PSA & ETS & UNI & SA\cite{fujii2016} & PSA & ETS & UNI & SA\cite{fujii2016} & PSA & ETS \\
    \midrule
    Maize, NAVAIL & 3.87 & 2.79 & 2.51 & 2.59 & 2.39 & 2.42 & 2.43 & 2.29 & 2.43 & 2.58 & 2.53 & 2.27 & 2.28 & 2.28 & 2.33 & 2.23 & 2.23 \\ 
    Maize, GRLV & 2.83 & 2.80 & 2.47 & 2.14 & 2.23 & 2.50 & 2.45 & 2.08 & 2.13 & 2.30 & 2.39 & 2.04 & 2.09 & 2.06 & 2.06 & 1.96 & 1.98 \\ 
    Millet, NAVAIL & 5.25 & 2.89 & 2.79 & 2.87 & 2.68 & 2.74 & 2.72 & 2.67 & 2.75 & 2.80 & 2.75 & 2.69 & 2.67 & 2.71 & 2.72 & 2.68 & 2.68 \\ 
    Millet, GRLV & 1.82 & 1.77 & 1.47 & 1.24 & 1.27 & 1.63 & 1.56 & 1.23 & 1.25 & 1.43 & 1.48 & 1.24 & 1.27 & 1.31 & 1.30 & 1.24 & 1.25 \\ 
    Sorghum, NAVAIL & 2.95 & 2.37 & 2.21 & 1.45 & 1.53 & 1.75 & 2.13 & 1.48 & 1.46 & 1.73 & 1.87 & 1.34 & 1.46 & 1.33 & 1.50 & 1.25 & 1.21 \\
    Sorghum, GRLV & 2.81 & 2.06 & 2.19 & 1.72 & 1.88 & 1.80 & 2.07 & 1.52 & 1.74 & 1.70 & 1.81 & 1.42 & 1.62 & 1.34 & 1.44 & 1.24 & 1.34 \\
    Wheat, NAVAIL & 4.03 & 2.49 & 2.14 & 1.96 & 1.96 & 2.05 & 1.96 & 1.85 & 2.08 & 2.16 & 2.08 & 1.82 & 1.97 & 1.91 & 1.93 & 1.87 & 1.82 \\
    Wheat, GRLV & 1.32 & 1.29 & 0.94 & 0.78 & 0.83 & 1.26 & 1.10 & 0.77 & 0.81 & 1.11 & 1.11 & 0.80 & 0.82 & 0.95 & 0.90 & 0.83 & 0.83 \\ 
    \bottomrule
  \end{tabular}
  \label{tbl:losses-cropsim}
\end{table*}

\subsection{Cold-Hardiness Prediction}
Cold temperatures can cause significant frost damage to fruit crops depending on their resilience, or cold hardiness, which changes throughout the dormancy season. Knowing this value is important for making management decisions regarding the deployment of expensive frost-mitigation measures. Unfortunately, cold hardiness is difficult and costly to measure and can fluctuate significantly from day to day in response to local weather patterns. Additionally, cold hardiness can vary significantly across grape cultivars. There are cultivar specific predictive models for cold hardiness \cite{c:ferguson2014,c:saxena2023}, but only for a small number of grape cultivars. For the many other grape cultivars, there is no clear way to get cultivar-specific predictions.

This work uses grape cold-hardiness data collected from 1988 to 2022 at the WSU Irrigated Agriculture Research and Extension Center in Prosser, WA for 18 grape cultivars. During this period grape bud samples were collected and analyzed approximately every two weeks during dormancy to determine the ground truth cold-hardiness values. The number of seasons of data collection vary across cultivars, ranging from 4 to 30. The dataset also includes daily weather data from the closest weather station, which is obtained from the AgWeatherNet API \cite{c:agnet}. The dataset is constructed using dormant season data from September 7th to May 15th. Seasons with more than 10\% of missing weather data are excluded. 

Since the data set only has ground truth values approximately every two weeks, it is difficult to conduct BOAL experiments using just this dataset. In order to simulate sampling any day and not be limited by the data sparsity, we fill the season labels using a single neural network trained for the target cultivar. This filling is only used for simulating the querying process of BOAL, and all evaluations of learned models are on the real-world data samples. The limited nature of the data means that only some cultivars have high accuracy models. To minimize artifacts from model error, we select the 8 cultivars with the highest accuracy models to use as the target function for defining 8 distinct BOAL problems. For BOAL experiments involving a particular target cultivar, we define a set of 17 experts derived from a multi-task cold-hardiness model \cite{c:saxena2023} trained on all cultivars excluding the target cultivar. The implementation details of this model are in Appendix \ref{sec:mtl-model}.

\subsection{Experimental Protocol}
\label{sec:protocol}

We evaluate several BOAL methods on both simulated agricultural problems and the real-world grape cold-hardiness problems. For each crop and prediction target pair, we conduct multiple BOAL experiments, each corresponding to one growing season for WOFOST or domancy period for cold-hardiness. In WOFOST simulations, we select one of the 15 models as the target and use the remaining 14 as experts. We report the average over all 15 targets. For grape cold hardiness, the target is one of 8 target cultivars, with models for the remaining 17 cultivars serving as experts. This setup simulates realistic scenarios where the target is not exactly captured by the experts, but at least one expert is expected to provide useful predictions due to the diversity of experts. 

We test multiple query budgets, ranging from 2 to 5 for cold hardiness and 2 to 10 for WOFOST, chosen to reflect realistic constraints in actual applications. For each target function and budget, we conduct 37 BOAL runs, each using different historical years of weather data. Performance is measured via Root Mean Square Error (RMSE) with respect to the target, with the average RMSE over the 37 runs reported as the final performance metric. For our new methods leveraging historical data, we use the other 36 years of weather data from seasons not present in the test set. This protocol ensures a comprehensive evaluation across various targets, budgets, and historical weather patterns, while maintaining a realistic simulation of the challenges faced in agricultural applications.

In all experiments, we run the \textsc{Hedge} algorithm with a learning rate of $\eta=1$. We evaluate four methods: 1) \textbf{Base} - uniformly weighted expert set with no queries, 2) \textbf{UNI} - temporally uniform query selection, 3) \textbf{SA} - the most closely related prior work \cite{fujii2016}, which uses the Secretary Algorithm (SA) for \textsc{OnlineMax} and does not use prior episodic data, 4) \textbf{PSA} - an instantiation of Algorithm \ref{alg:general} using the prophet-secretary algorithm, and 5) \textbf{ETS} - our empirical threshold selection algorithm used to instantiate Algorithm \ref{alg:general}.

\begin{table*}[t!]
  \caption{Comparison of RMSE of cold-hardiness predictions for the different sampling methods. The column labels denote 1) the number of samples, 2) the sampling method.}
  \small
  \centering
  \begin{tabular}{llllll|llll|llll|llll}
    \toprule
    Budget & \multicolumn{1}{c}{0} & \multicolumn{4}{c}{2} & \multicolumn{4}{c}{3} & \multicolumn{4}{c}{4} & \multicolumn{4}{c}{5}\\
    \cmidrule(r){2-2}
    \cmidrule(r){3-6}
    \cmidrule(r){7-10}
    \cmidrule(r){11-14}
    \cmidrule(r){15-18}
    Cultivar & Base & UNI & SA\cite{fujii2016} & PSA & ETS & UNI & SA\cite{fujii2016} & PSA & ETS & UNI & SA\cite{fujii2016} & PSA & ETS & UNI & SA\cite{fujii2016} & PSA & ETS \\
    \midrule
    Chardonnay & 1.35 & 1.40 & 1.35 & 1.37 & 1.37 & 1.39 & 1.36 & 1.33 & 1.35 & 1.38 & 1.36 & 1.33 & 1.34 & 1.35 & 1.35 & 1.33 & 1.33 \\
    Grenache & 1.57 & 1.48 & 1.48 & 1.43 & 1.45 & 1.47 & 1.46 & 1.43 & 1.43 & 1.45 & 1.48 & 1.42 & 1.42 & 1.43 & 1.48 & 1.41 & 1.44 \\
    Merlot & 1.52 & 1.63 & 1.55 & 1.52 & 1.53 & 1.55 & 1.53 & 1.50 & 1.52 & 1.54 & 1.52 & 1.51 & 1.50 & 1.52 & 1.52 & 1.50 & 1.51 \\
    Mourvedre & 2.00 & 2.13 & 1.77 & 1.79 & 1.76 & 1.79 & 1.76 & 1.72 & 1.73 & 1.78 & 1.74 & 1.73 & 1.74 & 1.76 & 1.74 & 1.74 & 1.75 \\
    Pinot Gris & 1.50 & 1.67 & 1.67 & 1.64 & 1.67 & 1.71 & 1.69 & 1.65 & 1.68 & 1.69 & 1.70 & 1.64 & 1.63 & 1.64 & 1.69 & 1.62 & 1.65 \\
    Sangiovese & 1.88 & 1.81 & 1.50 & 1.51 & 1.48 & 1.65 & 1.49 & 1.46 & 1.45 & 1.60 & 1.52 & 1.48 & 1.45 & 1.56 & 1.50 & 1.51 & 1.46 \\
    Syrah & 1.31 & 1.47 & 1.38 & 1.37 & 1.37 & 1.38 & 1.37 & 1.32 & 1.36 & 1.37 & 1.35 & 1.33 & 1.34 & 1.35 & 1.34 & 1.32 & 1.35 \\
    Viognier & 1.51 & 1.50 & 1.46 & 1.51 & 1.43 & 1.55 & 1.45 & 1.49 & 1.43 & 1.50 & 1.50 & 1.46 & 1.41 & 1.47 & 1.49 & 1.44 & 1.42 \\
    \midrule
    Mean & 1.58 & 1.64 & 1.52 & 1.52 & 1.51 & 1.56 & 1.51 & 1.49 & 1.49 & 1.54 & 1.52 & 1.49 & 1.48 & 1.51 & 1.51 & 1.49 & 1.49 \\
    \bottomrule
  \end{tabular}
  \label{tbl:losses-CH}
\end{table*}

\begin{table*}[th!]
  \caption{Comparison of RMSE of cold-hardiness predictions for the different sampling methods with a worse set of experts. The column labels denote 1) the number of samples, 2) the sampling method.}
  \small
  \centering
  \begin{tabular}{llllll|llll|llll|llll}
    \toprule
    Budget & \multicolumn{1}{c}{0} & \multicolumn{4}{c}{2} & \multicolumn{4}{c}{3} & \multicolumn{4}{c}{4} & \multicolumn{4}{c}{5}\\
    \cmidrule(r){2-2}
    \cmidrule(r){3-6}
    \cmidrule(r){7-10}
    \cmidrule(r){11-14}
    \cmidrule(r){15-18}
    Cultivar & Base & UNI & SA\cite{fujii2016} & PSA & ETS & UNI & SA\cite{fujii2016} & PSA & ETS & UNI & SA\cite{fujii2016} & PSA & ETS & UNI & SA\cite{fujii2016} & PSA & ETS \\
    \midrule
    Chardonnay & 1.75 & 1.74 & 1.67 & 1.71 & 1.69 & 1.67 & 1.66 & 1.62 & 1.65 & 1.65 & 1.63 & 1.63 & 1.62 & 1.63 & 1.63 & 1.65 & 1.63 \\
    Grenache & 1.77 & 1.68 & 1.74 & 1.77 & 1.69 & 1.71 & 1.70 & 1.69 & 1.64 & 1.65 & 1.72 & 1.62 & 1.63 & 1.65 & 1.69 & 1.67 & 1.61 \\
    Merlot & 1.74 & 1.86 & 1.75 & 1.74 & 1.72 & 1.79 & 1.72 & 1.71 & 1.71 & 1.73 & 1.75 & 1.7 & 1.71 & 1.72 & 1.73 & 1.69 & 1.71 \\
    Mourvedre & 1.87 & 2.01 & 1.89 & 1.78 & 1.77 & 1.82 & 1.77 & 1.79 & 1.75 & 1.85 & 1.86 & 1.74 & 1.75 & 1.81 & 1.81 & 1.76 & 1.74 \\
    Pinot Gris & 1.51 & 1.61 & 1.52 & 1.55 & 1.53 & 1.66 & 1.58 & 1.54 & 1.52 & 1.60 & 1.57 & 1.50 & 1.49 & 1.52 & 1.61 & 1.52 & 1.50 \\
    Sangiovese & 1.74 & 1.75 & 1.39 & 1.42 & 1.44 & 1.57 & 1.42 & 1.38 & 1.41 & 1.50 & 1.42 & 1.41 & 1.43 & 1.48 & 1.40 & 1.41 & 1.40 \\
    Syrah & 1.29 & 1.47 & 1.46 & 1.45 & 1.44 & 1.43 & 1.45 & 1.42 & 1.42 & 1.43 & 1.44 & 1.41 & 1.42 & 1.42 & 1.43 & 1.42 & 1.42 \\
    Viognier & 1.38 & 1.34 & 1.40 & 1.41 & 1.35 & 1.38 & 1.37 & 1.36 & 1.31 & 1.33 & 1.35 & 1.35 & 1.31 & 1.31 & 1.34 & 1.31 & 1.30 \\
    \midrule
    Mean & 1.63 & 1.68 & 1.60 & 1.60 & 1.58 & 1.63 & 1.59 & 1.56 & 1.55 & 1.59 & 1.59 & 1.54 & 1.54 & 1.57 & 1.58 & 1.55 & 1.54 \\
    \bottomrule
  \end{tabular}
  \label{tbl:losses-CH-E}
\end{table*}

\subsection{Agricultural Simulator Results}

\textbf{Overall Benefit of BOAL.}
Table \ref{tbl:losses-cropsim} gives the results for each WOFOST crop and target combination for various budgets and methods.
We see that across all budget sizes considered, all of the active learning methods usually outperformed UNI, achieving lower losses even with very few samples. This gives evidence that there is benefit to being more selective about when to issue queries. Notably, our ETS approach frequently has the lowest losses of all methods considered. Below, we use the paired Wilcoxon Signed-Rank Test \cite{wilcoxon} to assess the significance of differences between the methods. 

\textbf{Impact of Queries.}
We compare the results of all query methods to Base, which does not use any queries. 
Using the Wilcoxon Test with a significance level of 0.05, we found that every query method is significantly better than Base, for all problems and budgets. This benefit of using queries in all cases shows that the expert weighting method is able to effectively use the labels collected, even with very few labels. This is particularly notable since the initial performance of the expert ensemble without any labels is already non-trivial. 

\textbf{Impact of Selective Queries.}
We next compare the results of all informed querying methods (SA, PSA, ETS) to temporally uniform queries (UNI).
In all cases, the Wilcoxon test showed that PSA and ETS are better than UNI. In over half of the cases, SA was also better than UNI, but was worse or showed no significant difference for remaining cases. 
This clear improvement over UNI is evidence that the methods are able to more effectively use the query budget by conditioning on prior observations and labels. However, the SA results indicate that using informed sampling with no prior episodic information is not always sufficient.

\textbf{Impact of Episodic Prior Knowledge.}
Lastly, we compare the results of our methods that use episodic prior knowledge (PSA, ETS) to SA, which does not use this knowledge. 
In all but one case for PSA and three cases for ETS, the Wilcoxon signed rank test shows that our methods significantly improve over SA, demonstrating the benefit of using this extra information when available. 

\begin{table*}[t!]
  \caption{Comparison of average \textsc{Score} value for different sampling methods. Each method is evaluated on four crops and two labels. The column labels denote 1) the number of samples collected, 2) the sampling method used, either SA, PSA, ETS, or Max Oracle, where Max Oracle has the full season information and will always pick the sample with the highest \textsc{Score} value.}
  \small
  \centering
  \begin{tabular}{lllll|llll|llll|llll}
    \toprule
    Budget & \multicolumn{4}{c}{2} & \multicolumn{4}{c}{3} & \multicolumn{4}{c}{4} & \multicolumn{4}{c}{10}\\
    \cmidrule(r){2-5}
    \cmidrule(r){6-9}
    \cmidrule(r){10-13}
    \cmidrule(r){14-17}
    Crop, Label & SA\cite{fujii2016} & PSA & ETS & Max & SA\cite{fujii2016} & PSA & ETS & Max & SA\cite{fujii2016} & PSA & ETS & Max & SA\cite{fujii2016} & PSA & ETS & Max \\
    \midrule
    Maize, NAVAIL & 20.1 & 16.9 & 26.9 & 28.7 & 16.0 & 16.9 & 15.8 & 19.8 & 12.2 & 12.7 & 14.2 & 15.3 & 6.4 & 6.6 & 6.8 & 7.0 \\
    Maize, GRLV & 9.3 & 28.0 & 34.4 & 47.0 & 5.9 & 17.0 & 26.0 & 33.1 & 7.4 & 13.6 & 17.6 & 22.0 & 6.2 & 6.9 & 7.5 & 9.6 \\
    Millet, NAVAIL & 20.5 & 14.1 & 21.5 & 21.8 & 15.4 & 13.9 & 17.5 & 18.0 & 13.4 & 10.3 & 12.3 & 12.5 & 6.6 & 6.4 & 6.6 & 6.6 \\
    Millet, GRLV & 4.3 & 8.4 & 11.8 & 14.4 & 1.7 & 5.9 & 7.6 & 9.9 & 2.4 & 4.4 & 5.2 & 7.0 & 2.3 & 2.5 & 2.5 & 3.0 \\ 
    Sorghum, NAVAIL & 7.8 & 11.7 & 15.1 & 18.8 & 6.3 & 9.3 & 10.1 & 11.1 & 5.7 & 7.6 & 7.6 & 9.4 & 3.4 & 3.8 & 4.0 & 4.2 \\
    Sorghum, GRLV & 12.8 & 27.9 & 30.5 & 30.5 & 7.0 & 18.4 & 21.5 & 20.5 & 8.1 & 14.1 & 15.5 & 17.9 & 5.2 & 5.8 & 7.0 & 8.8 \\
    Wheat, NAVAIL & 19.9 & 13.3 & 26.0 & 27.0 & 15.9 & 14.0 & 15.1 & 17.1 & 13.1 & 10.1 & 14.4 & 15.0 & 6.7 & 6.3 & 6.9 & 7.0 \\
    Wheat, GRLV & 2.9 & 6.9 & 9.2 & 13.2 & 0.6 & 4.6 & 6.7 & 8.6 & 0.6 & 3.4 & 4.6 & 6.0 & 1.7 & 1.9 & 2.1 & 2.5 \\ 
    \bottomrule
  \end{tabular}
  \label{tbl:ALscore-sim}
\end{table*}

\begin{table*}[t!]
  \caption{Comparison of average \textsc{Score} value for different sampling methods for cold-hardiness prediction. The column labels denote 1) the number of samples collected, 2) the sampling method used, either SA, PSA, ETS, or Max Oracle, where Max Oracle has the full season information and will always pick the sample with the highest \textsc{Score} value.}
  \small
  \centering
  \begin{tabular}{lllll|llll|llll|llll}
    \toprule
    Budget & \multicolumn{4}{c}{2} & \multicolumn{4}{c}{3} & \multicolumn{4}{c}{4} & \multicolumn{4}{c}{5}\\
    \cmidrule(r){2-5}
    \cmidrule(r){6-9}
    \cmidrule(r){10-13}
    \cmidrule(r){14-17}
    Cultivar & SA\cite{fujii2016} & PSA & ETS & Max & SA\cite{fujii2016} & PSA & ETS & Max & SA\cite{fujii2016} & PSA & ETS & Max & SA\cite{fujii2016} & PSA & ETS & Max \\
    \midrule
    Chardonnay & 1.21 & 1.53 & 2.15 & 2.83 & 1.42 & 1.45 & 1.71 & 2.22 & 1.06 & 1.55 & 1.81 & 1.87 & 1.08 & 1.57 & 1.64 & 1.64 \\ 
    Grenache & 1.03 & 1.61 & 2.43 & 3.25 & 1.39 & 1.46 & 1.81 & 2.24 & 0.78 & 1.45 & 1.77 & 1.77 & 0.74 & 1.38 & 1.67 & 1.66 \\
    Merlot & 1.12 & 1.42 & 2.38 & 3.17 & 1.18 & 1.38 & 1.76 & 2.17 & 0.73 & 1.38 & 1.72 & 1.72 & 0.74 & 1.31 & 1.53 & 1.56 \\
    Mourvedre & 1.06 & 1.51 & 2.36 & 3.21 & 1.18 & 1.34 & 1.75 & 2.22 & 0.61 & 1.31 & 1.72 & 1.77 & 0.63 & 1.29 & 1.60 & 1.59 \\
    Pinot Gris & 1.39 & 2.08 & 2.41 & 3.00 & 1.79 & 1.81 & 1.95 & 2.57 & 1.32 & 1.93 & 2.19 & 2.28 & 1.31 & 1.94 & 2.03 & 2.09 \\
    Sangiovese & 1.11 & 1.60 & 2.47 & 3.47 & 1.35 & 1.45 & 1.79 & 2.33 & 0.72 & 1.40 & 1.74 & 1.77 & 0.70 & 1.51 & 1.68 & 1.63 \\
    Syrah & 1.06 & 1.50 & 2.41 & 3.22 & 1.24 & 1.36 & 1.82 & 2.26 & 0.62 & 1.38 & 1.74 & 1.79 & 0.61 & 1.35 & 1.65 & 1.62 \\
    Viognier & 1.15 & 1.71 & 2.72 & 3.23 & 1.30 & 1.54 & 2.04 & 2.57 & 0.73 & 1.58 & 2.03 & 2.06 & 0.81 & 1.47 & 1.86 & 1.83 \\
    \midrule
    Mean & 1.14 & 1.62 & 2.42 & 3.17 & 1.36 & 1.47 & 1.83 & 2.32 & 0.82 & 1.50 & 1.84 & 1.88 & 0.83 & 1.48 & 1.71 & 1.70 \\
    
    \bottomrule
  \end{tabular}
  \label{tbl:ALscore-CH}
\end{table*}

\subsection{Real-World Cold-Hardiness Results}

We repeat the same experiments for real-world grape cold-hardiness. Table \ref{tbl:losses-CH} shows the results of these experiments for each of the target cultivars. Like the WOFOST problems, we see a clear benefit from collecting informed samples and using episodic prior knowledge. The Wilcoxon signed-rank test shows that in all but one case (two sample UNI), every method outperformed Base. Similarly, all but one case of the informed methods (SA with 5 queries) beat UNI and all but one case of the episodic priors methods (PSA with two queries) beat SA. These improvements, while statistically significant, are much smaller and diminish quicker with an increase in budget compared to the WOFOST results. One reason for this is likely that the reported test RMSE loss in these experiments is based on real-world data, which contain non-trivial amounts of noise.

\textbf{Lower Quality Experts.} We also consider how these results change with a lower quality set of experts. Specifically, as described in Appendix \ref{sec:mtl-model}, our multi-task cold-hardiness model allows us to sample virtual cultivars from the learned cultivar-embedding space. This lets us randomly sample a set of 17 experts to use for BOAL. In general, we have found that these virtual cultivars tend to have lower overall performance compared to the first set of experts. 

Table \ref{tbl:losses-CH-E} gives the results for this new set of experts. The results of Base show that this new set of experts performs worse than the original set of experts (Table \ref{tbl:losses-CH}), but not so much worse that it would be unreasonable to use them. While we do see a worse performance for all methods, the relative performance between methods is similar. In all but one case (SA vs. UNI with 4 samples), the Wilcoxon test results show the same significant differences as observed for the original experts. This suggests that the methods are robust across different quality expert sets. 


\subsection{Investigating OnlineMax}



The above results showed that our new methods that leverage prior episodic data outperform methods that ignore that data. Here we aim to understand whether this improvement is correlated to the performance of \textsc{OnlineMax} or was due to some other unknown factors. Recall that the objective of \textsc{OnlineMax} when called for a particular time interval is to select a query that is close the maximum active-learning score (Equation \ref{eq:score}) across that interval. To measure the effectiveness of different variants of \textsc{OnlineMax}, Tables \ref{tbl:ALscore-sim} (WOFOST) and Table \ref{tbl:ALscore-CH} (cold hardiness) show the average score returned for each variant over all calls to \textsc{OnlineMax} in our experiments. We consider the three selective query methods (SA, PSA, ETS) as well as Max Oracle, which always selects the highest score in hindsight, representing an upper performance bound. 

We see that the average score for PSA and ETS are higher than SA, and ETS is almost always better than PSA, especially for smaller budgets. These relative differences are highly correlated with the earlier presented BOAL performance of these methods. Interestingly, we also see that ETS is often close to Max Oracle, suggesting that ETS is close to optimal for selecting the sample with the highest score value in this setting. Note that we also observe that the absolute value of scores tends to decrease for larger budget sizes. This is explained by the fact that as the expert weights are updated more, their weighted variance tends to decrease. Overall these results are strong evidence that the observed benefits of our PSA and ETS methods over the most closely related prior work SA is due to the incorporation of prior episodic knowledge. 

To better understand the poor relative performance of SA, we examined specific query selections and identified two common failure modes. First, in some cases, the maximum score occurs in the initial observational window of SA, which means SA will not select a query until the end of the interval, which can often have a significantly lower score than the max. Second, in other case, the observational window contains only small scores and then the scores rise quickly after that window. This leads SA to select the first score greater than those in the observational window, which again can be much smaller than the max. In both of these cases, SA is hindered by not being aware of the statistics of the input stream. The PSA and ETS methods are able to overcome both of those failure modes using their prior episodic knowledge.  


\section{Summary}
We considered how to approach active learning problems, where label querying is constrained by an online, finite-horizon setting, and a very small querying budget. We introduced a novel approach which integrates the learning-from-experts paradigm and prior unlabeled episodic data. Through experiments on both realistic simulated data and real-world agricultural data, we showed our method significantly outperforms baseline expert predictions, uniform query selection, and an existing BOAL method which does not take advantage of prior unlabeled episodic data. We are currently using our BOAL approach to collect four samples from four vineyards during the 2024-2025 dormancy period, with the goal of providing those vineyards with customized cold-hardiness models. In the future, we hope to identify and create additional realistic BOAL benchmarks for extremely small budget scenarios to encourage further progress in this area.

\newpage

\begin{acks}
This research was supported by NSF and USDA-NIFA under the AI Institute: Agricultural AI for Transforming Workforce and Decision Support (AgAID) award No.2021-67021-35344. The authors thank Lynn Mills and Alan Kawakami for the collection of cold-hardiness data.
\end{acks}

\bibliographystyle{ACM-Reference-Format}
\bibliography{base}

\appendix

\section{Cold-Hardiness Model Details}
\label{sec:mtl-model}

The multi-task cold-hardiness model is a neural network with four layers and an additional 12-dimensional embedding vector. This vector is cultivar specific and is concatenated with each time step of the input weather data before being fed into the neural network model. We can use this cultivar-specific embedding vector space to generate more expert models. To create these experts, we randomly generate cultivar embeddings by sampling a value for each embedding feature from a uniform distribution constrained by the maximum and minimum feature value in the original cultivar embeddings. 

The first layer of the multi-task model has 1024 fully connected nodes that output to a ReLU layer. The second layer has 2048 fully connected nodes that output to a ReLU layer. The third layer is a Gated Recurrent Unit (GRU) with a single layer of 2048 nodes. The fourth layer has 1024 fully connected nodes that output to a ReLU layer. The final layer is the prediction head to get the final prediction. For training, the Adam optimizer \cite{c:kingma2017} is used with a learning rate of 0.001. The batch size is 12, shuffled, and the model is trained for 400 epochs. The models are trained with the Binary Cross-Entropy of the phenology predictions at each time step and the Mean Square Error (MSE) of the cold-hardiness at time steps with ground-truth measurements available. The performance is evaluated using the Root Mean Square Error (RMSE) of the prediction.

\end{document}